\documentclass[11pt]{article}

\usepackage[final]{acl}

\usepackage{times}
\usepackage{latexsym}

\usepackage[T1]{fontenc}

\usepackage[utf8]{inputenc}

\usepackage{microtype}

\usepackage{inconsolata}

\usepackage{graphicx}
\usepackage{booktabs}
\usepackage{tabularx}
\usepackage{amssymb}
\usepackage{pifont}
\usepackage{placeins}
\usepackage{enumitem}

\newcommand{\cmark}{\ding{51}}
\newcommand{\xmark}{\ding{55}}
\newcommand{\na}{\textemdash}

\title{Exposure is Optional: Learning Unlike Coordination in Language Models}

\author{Jiamu Luo \\
  New York University\\
  \texttt{jiamu.l@nyu.edu} \\\And
  Shane Steinert-Threlkeld \\
  University of Washington \\
  \texttt{shanest@uw.edu} \\}

\begin{document}
\maketitle
\begin{abstract}
Coordination, a fundamental linguistic structure, remains a subject of intense debate. A common view holds that only same-category constituents can be conjoined, which has been challenged by the many grammatical unlike coordinations found in natural language. Treating language models as a computational testbed, we investigate whether the acquisition of unlike coordination requires direct exposure in the training data, or whether it can emerge organically from general compositional abilities. Using Filtered-Corpus Training (FiCT), we train GPT-2 models on corpora from which all instances of unlike coordination have been removed. We find that direct exposure is not necessary: models trained on filtered data successfully generalize to unlike coordination, achieving perplexity and grammaticality judgments comparable to models trained on unfiltered text. Furthermore, our analyses of internal representations indicate that language models process unlike coordination by treating the conjoined elements as belonging to similar structural categories or through a mechanism akin to deletion, both of which appear learnable from exposure to alike coordination alone.
\end{abstract}

\section{Introduction}

Coordination allows speakers to combine multiple constituents into one larger unit, with rough semantic implications that the conjuncts should be similar and equal in various perspectives. Theoretical treatments of coordination often assume that two elements may be coordinated only if they share the same syntactic category, a principle known as the Law of Coordination of Likes (LCL) \citep{chomskySyntacticStructures1957a, williamsAcrosstheBoardRuleApplication1978a, williamsTransformationlessGrammar1981a}. However, there are apparent counterexamples. Conjuncts can differ syntactically, as in \textit{you can depend on} [NP \textit{my assistant}] \textit{and} [CP \textit{that he will be on time}]. Such cases appear to have ``coordination of unlikes.” We leverage language models (LMs) as a computational lens to explore this phenomenon, asking whether models require direct exposure to learn unlike coordination, or if they can generalize it entirely from the structures of alike coordination.

Unlike human language acquisition, language models offer precise control over the training stimulus. To exploit this, we utilize Filtered-Corpus Training (FiCT), as termed by \citet{patilFilteredCorpusTraining2024a}, to establish a clean-slate learning environment where models are exclusively exposed to alike coordination. Specifically, we train models from scratch on  corpora from which unlike coordinations have been systematically filtered and compare their generalization behaviors and internal representations against the models trained on unfiltered text. This strict isolation allows us to test whether direct exposure to category mismatches in coordination is necessary for acquisition.

We demonstrate that unlike coordination is not a general exception to LMs and can be learned effectively from the indirect information present in alike coordination and other linguistic expressions. We show that LMs process unlike coordination via two distinct mechanisms: treating conjoined elements as belonging to a shared "supercategory," or through a process akin to deletion. Both mechanisms appear to be learnable from exposure to alike coordination alone. We then connect these computational behaviors directly to theoretical linguistic debates. The successful performance of the models suggests that it does not presuppose a constraint where conjuncts must be of identical syntactic categories. Crucially, this can be viewed as evidence that individually motivated mechanisms and specialized grammatical rules are theoretically unnecessary.

Moreover, our study highlights the value of linguistically grounded evaluations for interpreting model behavior. We curate a targeted suite of grammaticality judgment tests drawn from theoretical linguistics literature to probe the boundaries of model knowledge. These diagnostics expose nuanced error and generalization patterns on subtle syntactic constraints across different training condition.

\section{Background and Related Work}

\paragraph{Theoretical treatments of coordination}
A historically dominant perspective relies on the Law of Coordination of Likes (LCL), which posits that only elements of the same syntactic category can be coordinated \citep{chomskySyntacticStructures1957a, williamsAcrosstheBoardRuleApplication1978a, williamsTransformationlessGrammar1981a}. To account for grammatical counterexamples within this framework, theorists often reinterpret unlike conjuncts to achieve underlying category matching. Proposed mechanisms include analyzing categories as feature bundles with shared supersets \citep{sagCoordinationHowDistinguish1985a}, or postulating abstract ''supercategories``, such as PredicateP \citep{bowersSyntaxPredication1993a} and ModifierP \citep{rubinModificationSyntacticAnalysis1994a}, alongside invisible elided structures to unify the mismatched conjuncts \citep{brueningCategoryMismatchesCoordination2020a}.

A competing line of inquiry rejects the necessity of the LCL. These approaches treat conjunctions as heads of their own phrases, allowing diverse conjuncts to occupy specifier, complement, or adjunct positions without mandating category identity \citep{johannessenCoordination1998a, munnNULLOPERATORANALYSIS1992a, kayneAntisymmetrySyntax1994a}. \citet{patejukCategoryMismatchesCoordination2023a} argue that the LCL is effectively illusory and that complex workarounds like ellipsis or null heads are overly abstract.

\paragraph{Assessing linguistic knowledge in language models}
Studies of linguistics knowledge in language models have grown from simple diagnostic classifiers \citep[][i.a.]{hupkesVisualisationDiagnosticClassifiers2018a, giulianelliHoodUsingDiagnostic2018a, belinkovAnalysisMethodsNeural2019a} to sophisticated structural probes and successfully uncover a hierarchy of linguistic knowledge, from surface-level phenomena like word length or morphological features, to deeper semantic and relational structure, including long-range dependencies and world-knowledge facts \citep[][i.a.]{conneauWhatYouCan2018b, liuLinguisticKnowledgeTransferability2019a, sahinLINSPECTORMultilingualProbing2020a, klafkaSpyingYourNeighbors2020a, hewittStructuralProbeFinding2019a}. Shifting from internal representations to external behavior, targeted syntactic evaluations \citep[][i.a.]{linzenAssessingAbilityLSTMs2016a, jumeletLanguageModelsUnderstand2018, marvinTargetedSyntacticEvaluation2018a, kannVerbArgumentStructure2019} offer a complementary methodology, first widely popularized for LSTMs by \citet{linzenAssessingAbilityLSTMs2016a}. This approach adopts the minimal-pair paradigm, borrowed from experimental linguistics, to assess the linguistic knowledge of language models. Researchers have tested a wide array of complex syntactic structures, including argument structure, island constraints, and negative polarity item licensing \citep[][i.a.]{marvinTargetedSyntacticEvaluation2018a, warstadtNeuralNetworkAcceptability2019a, futrellNeuralLanguageModels2019}. By directly testing models on these targeted challenges, this approach provides a clear behavioral measurement of a model’s linguistic competence.

\paragraph{Controlled training environments}\label{fict}
Unlike humans, artificial language learners allow precise manipulation of their training environments. A prominent methodology leveraging this advantage is Filtered-Corpus Training (FiCT) \citep{warstadtWhatArtificialNeural2022a,patilFilteredCorpusTraining2024a, misra-mahowald-2024-language, LEONG2026104751}, which involves training a model from scratch on a dataset that has been intentionally and systematically altered. As \citet{warstadtWhatArtificialNeural2022a} emphasizes, this approach enables precise and hypothesis-driven investigations into the mechanisms of language learning in neural networks. By evaluating models trained on corpora systematically stripped of specific linguistic structures, FiCT isolates the minimal and sufficient conditions required for acquisition.
\section{Methods}

The code and data for this study has been uploaded on GitHub, and will be provided if accepted.

\subsection{Overview: Filtered-Corpus Training (FiCT)}
Filtered-Corpus Training (FiCT) is the core methodology of this research. The central logic involves training multiple models from scratch on two distinct types of corpora and evaluating them on curated test sets: 1. A baseline corpus, which is an original and unfiltered collection of text. 2. One or more filtered corpora, from which some text is removed according to some criteria. In this study, we remove occurrences of unlike coordination in two variants: all unlike coordination, and unlike coordination connected specifically by \textit{and} (unlike conjunction).

\subsection{Data}
The training data are derived from the training corpus of the English corpus used and released by \citet{gulordavaColorlessGreenRecurrent2018a}. We also use their vocabulary for tokenizers. To obtain filtered variants, we apply two filtering procedures to the base corpus. Filter I produces the corpus \textit{filtered-all}, which excludes all sentences containing unlike coordination. Filter II produces the corpus \textit{filtered-and}, which excludes all sentences containing unlike conjunctions.

We further downsample the base corpus and \textit{filtered-and} to match the number of sentences in \textit{filtered-all}, yielding three training corpora: \textit{original}, \textit{filtered-and}, and \textit{filtered-all}. Each corpus consists of 1,948,104 sentences, though the total number of tokens varies. Using token counts measured with a GPT-2 tokenizer, \textit{filtered-and} retains 91.02\% of the tokens in \textit{original}, while \textit{filtered-all} retains 88.49\%.

\subsection{Filters}
The filters are built on top of the Berkeley Neural Parser \citep{kitaevConstituencyParsingSelfAttentive2018a}, and use the model \texttt{benepar\_en3} to parse each sentence in the training corpus. Coordination structures are automatically identified based on the presence of the CC (coordinating conjunction) label. Sentences containing unlike coordination are removed according to the criteria defined by each filter type: filter I simply removes them, and filter II removes them if the node labeled CC is the word \textit{and}.

Since no established benchmark currently exists for evaluating such filters on this topic, we assess their performance using our own evaluation sets. The evaluation sets for the filter performance contain a set of 100 sentences of unlike coordination (\textit{unlike}) and a set of the corresponding 100 sentences of alike coordination (\textit{alike-from-unlike}). Details of the two evaluation sets are shown in Section~\ref{evaluation}. We report filtering results in Table~\ref{tab:filter-quality}.

\begin{table}[t]
\centering
\small
\begin{tabularx}{\columnwidth}{l *{4}{>{\centering\arraybackslash}X}}
\toprule
\textbf{FILTERS} & \textit{u-all} & \textit{u-wrong} & \textit{a-all} & \textit{a-kept} \\
\midrule
Filter I  & 100 & 1 & 100 & 83 \\
Filter II & 100 & 1 & 100 & 83 \\
\bottomrule
\end{tabularx}
\caption{Filtering results on the evaluation sets. 
\textit{u} = unlike set, \textit{a} = alike set; 
\textit{all} = total sentences, \textit{wrong} = wrongly retained, \textit{kept} = correctly kept.}
\label{tab:filter-quality}
\end{table}

As suggested by the evaluation results, these two filters are strong filters. Each filter wrongly retains only a single sentence in \textit{unlike} due to parsing errors, but also filters away some sentences in \textit{alike}, which ideally should remain. Closer investigation reveals that these wrongly filtered sentences are because of parsing error, the complex case of \textit{be}, and the strict implementation of the filters, which discard conjoined phrases that are not labeled identically, even if they might actually belong to the same category by human judgment. This kind of strong filters may be, in fact, favorable. While additional removal of alike cases could reduce data representativeness and potentially degrade the performance, making the evaluation results less convincing, our results suggest that this is not the case. On the other hand, the filters demonstrate high precision, effectively eliminating undesired unlike coordination. Consequently, these filters produce high-quality filtered corpora, providing a solid foundation for further analysis.

\subsection{Model training}
We use a causal, decoder-only Transformer \citep{Vaswani+2017} based on the GPT framework \citep{radfordLanguageModelsAre2019a}, with a configuration comparable to GPT-2 Large (1024 context length, 1280 embedding size, 36 layers, and 20 heads).

We trained three models variants: \textit{original}, \textit{filtered-and}, and \textit{filtered-all}, each on its corresponding training corpus of the same name. We implemented these models using the Huggingface \texttt{transformers} package \citep{wolf-etal-2020-transformers}. To account for variability introduced by initialization, we trained multiple models with different random seeds for each variant. We used a random number generator to generate three random numbers: 291, 1543, and 9071, which serve as the random seeds. So, in total we trained 9 models, 3 models of different random seeds for each variant. Please see Appendix~\ref{chap:training details} for the details of model training.

\subsection{Evaluation}\label{evaluation}
\subsubsection{Evaluation corpora}
We manually constructed five evaluation corpora, based on relevant literature or adapted from examples of unlike coordination in the Penn Treebank \citep{marcus-etal-1993-building} found by \citet{kalliniCorpusbasedSyntacticAnalysis2021a}: \textit{unlike}, \textit{alike-from-unlike}, \textit{supercat-deletion}, \textit{deletion}, and \textit{unlike-judgment}. Each corpus targets a specific aspect of model evaluation. Appendix~\ref{chap:examples-eval-corpora} provides examples from the evaluation corpus. The complete set of sentences are in our repository.

\textit{Unlike}: This evaluation corpus contains 100 sentences that have unlike coordination. Following the findings in \citet{kalliniCorpusbasedSyntacticAnalysis2021a} on the ten most frequent unlike coordination patterns and their relative frequencies, the corpus includes instances of: NP + SBAR, NP + VP, AP + VP, AdvP + PP, NP + AP, PP + VP, PP + AdvP, NP + PP, PP + NP, and VP + NP. The number of sentences for each pattern was adjusted to approximate the relative frequencies, resulting in the following distribution: 18, 15, 11, 10, 10, 9, 9, 7, 6, and 5 sentences, respectively. The sentences are drawn from relevant literature when exemplifying a given pattern, or from the Corpus of Contemporary American English \citep{DVN/AMUDUW_2015} presented in \citet{kalliniCorpusbasedSyntacticAnalysis2021a}. When these sources did not provide sufficient examples for a given pattern, additional sentences were constructed based on attested examples from these sources.

\textit{Alike-from-unlike}: This evaluation corpus consists of 100 sentences that have alike coordination, created on the basis of the sentences in \textit{unlike}. Each sentence was adapted from a corresponding sentence in \textit{unlike} with only minor modifications. This evaluation corpus can thus be viewed as a loose minimal-pair counterpart to \textit{unlike}.

\textit{Supercat-deletion}: This evaluation corpus consists of sentences from \textit{unlike} that show clear evidence of presupposing a supercategory, and they may also be interpreted as involving the deletion of shared elements. For example, in \textit{The phenomenon fall into place organically and with ease}, we can posit a shared supercategory of modifier phrases spanning both conjuncts. Alternatively, it may be viewed as derived from the deletion of the repeated phrase \textit{the phenomenon fall into place} at the beginning of the second conjunct. In total, the corpus contains 62 sentences.

\textit{Deletion}: This evaluation corpus consists of sentences from \textit{unlike} that can be interpreted as involving deletion, but are incompatible with the same-supercategory hypothesis. For example, \textit{The movie was funny and made everyone laugh} can be interpreted as involving deletion of the shared subject \textit{the movie}; however, it is difficult to identify a non-overgenerating supercategory that accounts for this coordination. In total, this evaluation corpus contains 24 sentences.

\textit{Unlike-judgment}: This evaluation corpus consists of 22 grammaticality judgment tests related to unlike coordination. The sentences were collected from relevant literature on coordination \citep{brueningCategoryMismatchesCoordination2020a, patejukCategoryMismatchesCoordination2023a, sagCoordinationHowDistinguish1985a, zhangCoordinationSyntax2009a, petersonProblemsConstraintsCoordination1981a,grosuSUBCATEGORIZATIONPARALLELISM1985a} and adapted for this study and context. The tests are designed in a loose minimal-pair pattern. Note that two tests contain only grammatical sentences. These are challenging cases involving unlike supercategories or single phrases with dual category status; in such cases, we focus on whether the models exhibit large differences in judgment rather than whether they can reject ungrammatical ones.

\subsubsection{Evaluation metrics}
\paragraph{Perplexity:}
This is used to evaluate the general performance of models in the evaluation corpora excluding \textit{unlike-judgment}.
\paragraph{Average surprisal:}
Average surprisal, as the logarithmic form of perplexity, measures how unlikely a sentence is under a model, and is used to determine whether the model prefers grammatical sentences over ungrammatical ones in \textit{unlike-judgment}. Passing a test means the model assigns a higher average surprisal to the ungrammatical sentence than to the grammatical one.
\paragraph{Average attention score:}
Average attention score is defined as \[\bar{A}(q, K) = \frac{1}{L \cdot H \cdot |K|} \sum_{\ell=1}^{L} \sum_{h=1}^{H} \sum_{k \in K} A^{(\ell, h)}_{q, k}\] where $\bar{A}(q, K)$ denotes the average attention from a query token $q$ to a key phrase $K$, $L$ is the number of Transformer layers considered, $H$ is the number of attention heads, $|K|$ is the number of tokens in the key phrase, and $A^{(\ell, h)}_{q, k}$ represents the attention weight from query token $q$ to key token $k$ in layer $\ell$ and head $h$. In this study, the score is computed by averaging attention values from the middle layer to the final layer to capture all relevant contextual information involved in processing and forming the representation of the query token.

Averaging attention score provides a summary signal to measure the overall degree to which the model allocates attention from the query token to the key phrase across layers, heads, and phrase tokens. Another common method in studying attention is to use the maximum, which captures the strongest individual attention link, but can also be sensitive to spikes that might arise from irrelevant reasons. Since this study is concerned with the general attention patterns rather than identifying specialized heads, the average is the more appropriate summary statistic characterizing the overall attention allocated to the phrase of interest.

To compare and decide if models have different processing patterns for theoretically different types of unlike coordination, for the sentences in \textit{deletion} and an equal number of sentences in \textit{supercat-deletion}, the average attention score is used to investigate how the models process these sentences. Given a query word, specifically, the first word in the second conjunct of the sentences in \textit{deletion} and \textit{supercat-deletion}, this score can be interpreted as
    \begin{enumerate}
        \item The model has a more supercategory-based interpretation of generating the coordination if the average attention score from the query to the first conjunct is greater than the score to the phrase that occupies a potential deletion position.
        \item The model has a more deletion-based interpretation of generating the coordination if the opposite is true (excluding cases of equality).
        \item Undefined if an equality appears.
    \end{enumerate}
    For example, in the sentence \textit{She mentioned the project and that it was already approved}, the query word is \textit{that}, and the two key phrases of interest are the first conjunct \textit{the project} and the phrase occupying the potential deletion position \textit{She mentioned}. We compare the average attention scores from the query word to the two key phrases to determine whether the coordination is better explained by a supercategory-based analysis or by a deletion-based analysis.

\paragraph{Cosine similarity:}
For sentences in \textit{deletion} and an equal number of sentences in \textit{supercat-deletion} and \textit{alike-from-unlike}, the cosine similarity score is computed between the conjuncts of each sentence. They contrast with each other such as the deletion-based analysis should indicate the lowest similarity between conjuncts among the three for sentences in \textit{deletion}, and the supercategory-based account predicts increased similarity between conjuncts for sentences in \textit{supercat-deletion}, and \textit{alike-from-unlike} condition serves as a sanity check in which conjunct similarity should be the highest. For each corpus, we then calculate the mean cosine similarity across all sentences. This measure reflects whether the model tends to represent phrases similarly when they appear in coordination.

\paragraph{Clustering:}
The last-layer representations of the conjuncts are also used for a clustering analysis with the $k$-means algorithm, and $k=3$. For each model variant, we extract its last-layer representations of the conjunct phrases in a sentence in a corpus, and average the representations if needed, then use these vectors as input. For each of the three model variants, we train a clustering system on \textit{deletion}, together with an equal number of sentences from \textit{supercat-deletion} and \textit{alike-from-unlike}, in the way described above. This results in a total of $3 \times 3 \times 3$ clustering systems (for each model variant there are three random seed models, each trained on the three corpora described above, and three model variants in total). The clustering results are intended to reflect the representations of conjuncts.

\section{Results and Analysis}

\begin{figure}[t]
    \centering
    \includegraphics[width=1\linewidth]{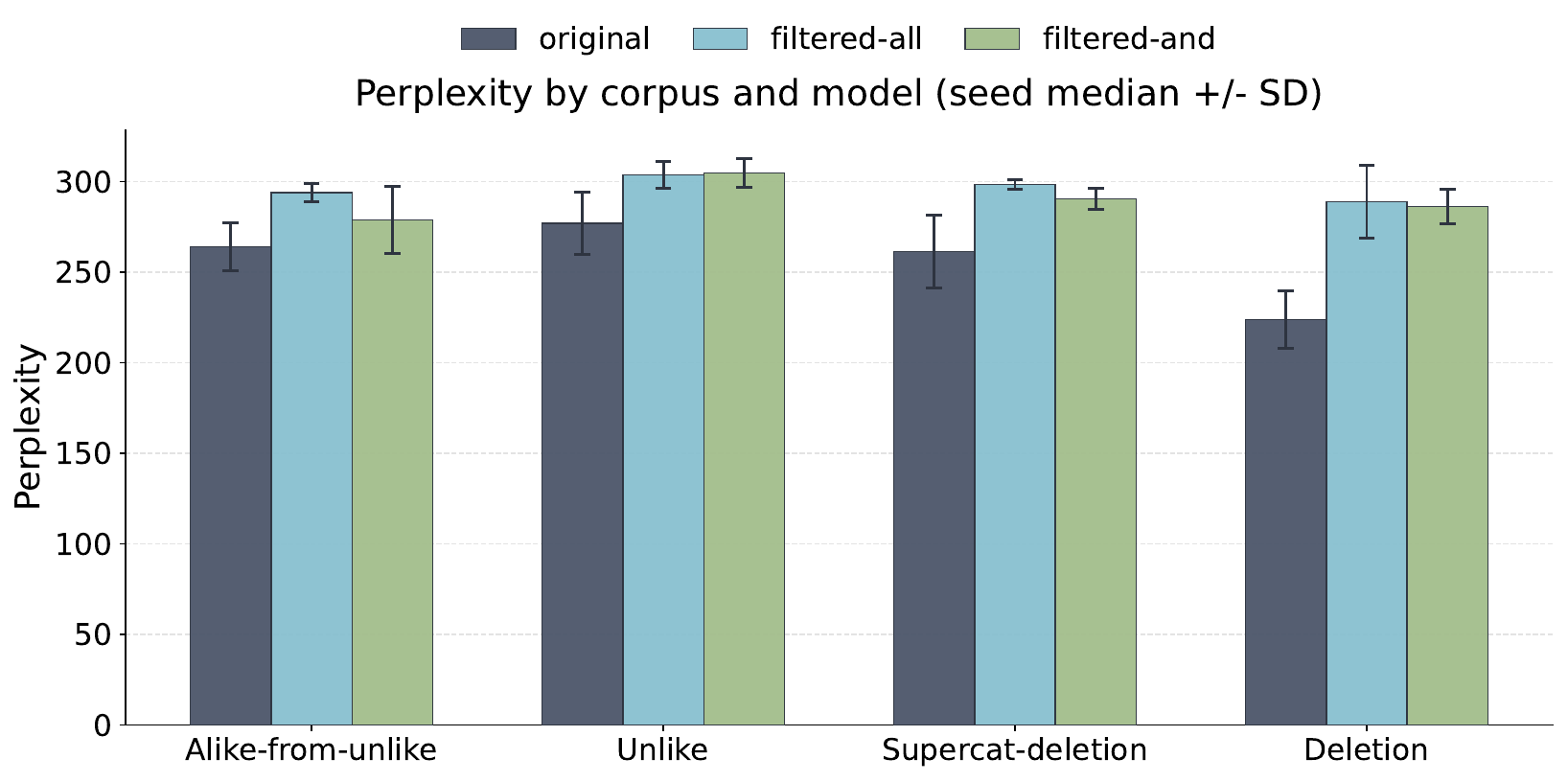}
    \caption{Mean perplexity of models on the evaluation corpora, excluding \textit{unlike-judgment}. The bar represents the seed median value.}
    \label{fig:model-perplexity}
\end{figure}

\paragraph{General performance}
Figure~\ref{fig:model-perplexity} summarizes the mean perplexity of each model across the evaluation corpora. The three model variants do not show a clear preference for either alike or unlike coordination, and a performance ordering of \textit{original} $>$ \textit{filtered-and} $>=$ \textit{filtered-all}. Also, the numerical advantage of the \textit{original} model over the filtered models is relatively modest.

In our pairwise \textit{t}-test results comparing all model variants, the \textit{original} model significantly outperforms both the \textit{filtered-all} and \textit{filtered-and} variants across all evaluation corpora. Conversely, there is no statistically significant difference between the two filtered models, despite a slight numerical advantage for the \textit{filtered-and} variant. Furthermore, our within-model analysis reveals no significant difference in perplexity between alike and unlike coordination for any model variant. In other words, once a model is fixed, its perplexity remains stable across coordination types, suggesting that any observed differences in the between-model analysis stem from model variation rather than inherent sensitivity to coordination type within a model.

\paragraph{Grammaticality judgment}

Table~\ref{tab:unlike-gj-summary} summarizes the results, with per-test details in Appendix~\ref{chap: Grammaticality Judgment Results}. Overall, models demonstrate a moderate ability to capture unlike coordination constraints, with the \textit{original} model achieving the highest accuracy. Also, there is no meaningful difference in performance among the three variants, which indicates that the penalty of filtering seems small for grammaticality judgment. The standard deviation further reveals a systematic effect of filtering on the stability of learned representations. Both the \textit{filtered-all} model and the \textit{filtered-and} model exhibit high stability, suggesting that reducing or eliminating conflicting evidence leads to more uniform and readily generalization. In contrast, the \textit{original} model shows moderate variability due to competing structural patterns in natural data.

\begin{table*}[t]
    \centering
    \begin{tabularx}{\textwidth}{
        >{\raggedright\arraybackslash}X
        >{\centering\arraybackslash}X
        >{\centering\arraybackslash}X
        >{\centering\arraybackslash}X
    }
    \toprule
     & \textit{Original} & \textit{Filtered-all} & \textit{Filtered-and} \\
    \midrule
    Median Row Acc.
        & 0.75 
        & 0.70 
        & 0.70 \\
    Mean Acc.\ ($\pm$ SD) & 0.70 $\pm$ 0.04 & 0.67 $\pm$ 0.02 & 0.65 $\pm$ 0.08\\
    Seed Corr.
        & 0.71
        & 0.79
        & 0.78 \\
    \bottomrule
    \end{tabularx}
    \caption{Summary statistics for the grammaticality judgment tests. The second row reports the median row accuracy, computed by comparing the median average surprisal assigned by a model to grammatical and ungrammatical sentences. The third row shows the mean accuracy across different random seeds, together with the corresponding standard deviation. The final row reports the mean Pearson correlation across random seeds. All values are rounded to two decimal places.}
    \label{tab:unlike-gj-summary}
\end{table*}

Across conditions, the models perform well on constraints that operate primarily at the surface-syntactic level, including verb dependency, deletion, and conjunct position effects. For example, they correctly identify that the unlike coordination is sensitive to the verb: \textit{...depend on my assistant and that he will be on time} is grammatical, but \textit{...strengthens the claim and that he will be on time} is infelicitous. They also recognize that certain unlike coordination constructions involve deletion, so \textit{She eats beans and with her hands} is acceptable, and they correctly dislike the counterpart that cancels out the deletion, as in \textit{Beans and with her hands, she eats}.

In contrast, the models show difficulty with cases in which category mismatch is governed by more fine-grained constraints. For instances, the models do not clearly prefer \textit{Hobbs ended up liking Rhodes and hating Barnes} over \textit{Hobbs ended up liking Rhodes and to hate Barnes}, and they prefers the wrong sentence \textit{You can depend on that he will be on time} instead of the correct sentence \textit{That he will be on time, you can depend on}, indicating limited sensitivity to restrictions on category mismatch. Overall, the models exhibit a bias of processing unlike coordination the same as alike coordination. The \textit{original} variant, trained on unfiltered natural data, can be even more generous.

Comparison of the error and success patterns across variants further suggests that all models develop a shared tendency toward organizing coordinated elements according to broad supercategorical similarities, which enables them to process many instances of unlike coordination. This tendency is supported positively by the models’ success on cases in which unlike conjuncts share a plausible supercategory, and negatively by their failure on more nuanced constraints where such constituents nevertheless cannot be conjoined. Moreover, in a test for unlike supercategory, all model variants exhibit a similar pattern, strongly disfavoring \textit{Reluctantly and embarrassed...}, relative to \textit{Reluctantly and in embarrassment...}. In other words, the models uniformly prefer cases in which the conjoined phrases can plausibly be analyzed as sharing a higher-level supercategory, while rejecting constructions that involve coordination across genuinely mismatched supercategories. As filtering decreases, the models appear to have an increasing capability to license deletion-based analyses. While this flexibility improves performance in some cases, it can also lead to overly permissive generalizations.

\paragraph{Attention pattern}

Table~\ref{tab:decision-attention-score} provides a concise summary of the results. We analyze the attention score differences using within-model Welch’s \textit{t}-tests and between-model paired \textit{t}-tests. Within models, attention patterns differ between \textit{deletion} and \textit{supercat-deletion} for all variants, though this difference is only statistically significant for the \textit{original} and \textit{filtered-and} models. Between models, processing of \textit{supercat-deletion} shows no significant differences. In contrast, for the \textit{deletion} condition, the \textit{original} model differs significantly from both filtered variants, while the difference between the filtered variants (no versus partial exposure) is nearly significant.

Overall, the models exhibit different attention patterns for \textit{deletion} versus \textit{supercat-deletion} sentences. Models exposed to unlike coordination show a larger distinction in how they process deletion-type versus supercategory-type unlike coordination, whereas the \textit{filtered-all} model exhibits a smaller difference. This reveals that the models can employ a different strategy to process unlike coordination in the \textit{deletion} condition, but require more exposure to fully develop and robustly deploy it.
\begin{table*}[t]
\centering
\small
\begin{tabularx}{\textwidth}{l *{4}{>{\centering\arraybackslash}X}}
\toprule
Model &
\multicolumn{2}{c}{\textit{Deletion}} &
\multicolumn{2}{c}{\textit{Supercat-deletion}} \\ 
\cmidrule(lr){2-3}
\cmidrule(lr){4-5}
& deletion-based & supercat-based & deletion-based & supercat-based \\
\midrule
\textit{Original} & 15.00 ± 1.41 & 9.00 ± 1.41 & 10.00 ± 0.47 & 14.00 ± 0.47\\
\textit{Filtered-all} & 12.00 ± 1.25 & 12.00 ± 1.25 & 10.00 ± 2.45 & 14.00 ± 2.45 \\
\textit{Filtered-and} & 14.00 ± 0.82 & 10.00 ± 0.82 & 10.00 ± 1.25 & 14.00 ± 1.25 \\
\bottomrule
\end{tabularx}
\caption{Overall results from the attention scores of a query word to the two key phrases on the two evaluation corpora. Numbers are median of seeds with the standard deviation.}
\label{tab:decision-attention-score}
\end{table*}

\paragraph{Similarity pattern}
Table~\ref{tab:average-similarity} reports the average similarity between conjuncts, along with the standard deviation, for each model across the evaluation corpora. The similarity scores clearly indicate that the models treat the three corpora differently: conjuncts in \textit{alike} show the highest similarity, those in \textit{supercat-deletion} exhibit moderate similarity, and those in \textit{deletion} show the lowest similarity.

\begin{table*}[t]
    \centering
    \begin{tabularx}{\textwidth}{*{4}{>{\raggedright\arraybackslash}X}}
    \toprule
        Model/Test Corpus& \textit{Alike-from-unlike} & \textit{Supercat-deletion} & \textit{Deletion}\\
    \midrule
        \textit{Original}&$0.6573 \pm 0.0027$ & $0.4880 \pm0.0086$ & $0.4002 \pm 0.0165$ \\
        \textit{Filtered-all}&$0.6512 \pm 0.0200$ & $0.4716 \pm 0.0159$ & $0.4264 \pm 0.0123$\\
        \textit{Filtered-and}&$0.6925 \pm 0.0198$ & $0.5113 \pm 0.0117$ & $0.4402 \pm 0.0325$\\
    \bottomrule
    \end{tabularx}
    \caption{Median value of average similarity for each seed of each model variant on each corpus with standard deviation across seeds.}
    \label{tab:average-similarity}
\end{table*}

We use a linear mixed-effects model to predict similarity from model variant, corpus, and their interaction (reference levels: \textit{original} and \textit{supercat-deletion}). Relative to \textit{supercat-deletion}, the \textit{alike} corpus yields significantly higher similarity ($\beta = 0.173$, $p < .001$), while \textit{deletion} significantly reduces it ($\beta = -0.084$, $p < .001$). The similarity increase for \textit{alike} is stable across all models. However, \textit{filtered-all} shows a significant positive interaction with \textit{deletion} ($\beta = 0.039$, $p = .002$), indicating that it experiences a smaller reduction in similarity compared to the \textit{original} model. No such interaction occurs for \textit{filtered-and} ($p = .987$). Overall, these results suggest that \textit{alike} produces the highest similarity scores, \textit{deletion} reduces similarity, and \textit{filtered-all} is specifically less affected by the deletion manipulation.

\paragraph{Clustering}
\begin{figure*}[ht]
    \centering
    \includegraphics[width=1\linewidth]{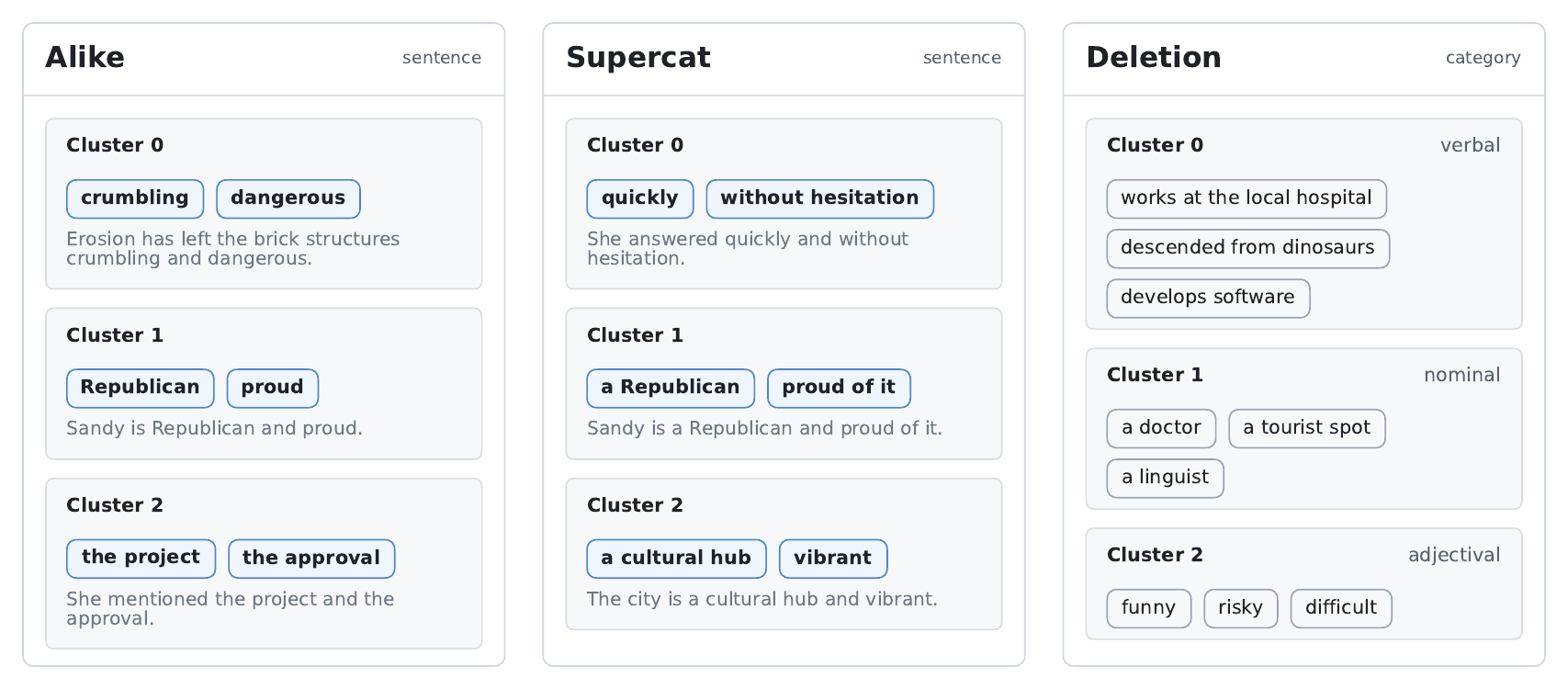}
    \caption{Illustration of clustering results. For each corpus, nine clustering systems are trained with different random seeds and different model variants. As they produce largely similar clusterings, we collapse them into a single representative system for visualization purposes.}
    \label{fig:cluster}
\end{figure*}

Figure~\ref{fig:cluster} provides an illustration of the clustering patterns. The clustering further reveals that the models treat both alike coordination in \textit{alike-from-unlike} and unlike coordination that can be considered alike within the same supercategory in \textit{supercat-deletion} as alike coordination, since phrases grouped together are mostly conjuncts from the same sentence. In contrast, for sentences in \textit{deletion}, where it is difficult to find an overarching alike category, the models recognize unlike coordination. Here, conjuncts from different sentences but sharing the same grammatical category are grouped together, showing a clear structure: nouns cluster with nouns, verbs with verbs, adjectives with adjectives, and so on.

\section{Discussion and Conclusion}

The first thing our findings demonstrate is that unlike coordination is not an exception or a special case for language models (LMs). Crucially, direct training exposure is not strictly necessary to process it: our filtered-all model generalized effectively on both perplexity and grammaticality judgments despite having all unlike coordination removed from its training data. This implies LMs lack an inherent inductive bias toward a strict Law of Coordination of Likes (LCL); had they presupposed such bias like the LCL, the filtered-all model would have catastrophically failed.

Instead, the ability to process unlike coordination emerges from general compositional capabilities learned via alike coordination. While latent structural cues (like supercategory similarity) are sufficient for models to license basic unlike coordination, direct exposure does refine performance. Models with exposure (\textit{original}, \textit{filtered-and}) are better equipped to exploit deletion-based mechanisms and accommodate nuanced structures like dual-role phrases. However, all models strongly reject combinations of unlike supercategories, suggesting that models rely heavily on conjunct similarity and struggle to bridge entirely opposing categories without specific distributional evidence.

Beyond the question of \textit{if} models learn unlike coordination, our internal representation analysis illuminates \textit{how} they process it. The models use two distinct mechanisms: supercategory and deletion. For supercategory cases, models cluster conjuncts by sentence context rather than syntactic category. This implies that the model constructs a unified functional "supercategory," deriving likeness from shared contextual function. In contrast, unlike coordination with deletion yields the lowest similarity scores and is clustered strictly by grammatical category across sentences. Attention patterns corroborate this distinction: in deletion cases, models treat conjuncts as structurally independent, with the second conjunct attending heavily to the subject rather than the first conjunct. While the \textit{filtered-all} model can utilize both strategies, exposure to unlike coordination significantly sharpens an LM's ability to cleanly distinguish between structures requiring a unified representation and those relying on implicit deletion.

For the theory of unlike coordination and coordination in general, LMs offer a perspective that challenges the necessity of the LCL. The models' dual processing strategies support a hybrid theoretical framework. The processing mechanisms align with the "Feature Bundle" and Supercategory theories \citep[][i.a.]{sagCoordinationHowDistinguish1985a, brueningCategoryMismatchesCoordination2020a} when a reasonable high-level node can be proposed. Meanwhile, cases that appear to involve deletion are best accounted for as true deletion constructions, rather than by appealing to other mechanisms. Most importantly, the fact that LMs are able to generalize from alike coordination to unlike coordination suggests that unlike coordination does not require a separately motivated grammatical mechanism. Consequently, theoretical accounts that posit individually-motivated constraints or mechanisms for unlike coordination may be unnecessarily strong. Unlike coordination appears to emerge naturally from the same representational resources that underlie coordination more broadly, together with general strategies, such as category underspecification, or deletion-like operations.



\section*{Limitations}

Our conclusions depend on the reliability of the parser-based filtering procedure. Although both filters removed 99 of the 100 unlike-coordination examples in our evaluation set, this does not establish that the filtered training corpora contain no unlike coordination. Parsing errors and constructions outside the evaluation set may leave residual examples. The filters also removed 17 of the 100 alike-coordination examples, indicating that filtering affects structures beyond the intended target. Our findings therefore support generalization under substantially reduced direct exposure, while the stronger claim of learning without any direct exposure requires further corpus validation.

The proposed supercategory-based and deletion-based interpretations rely on observational analyses of attention, cosine similarity, and clustering. These patterns are compatible with the proposed mechanisms, but do not uniquely identify them: shared context, lexical content, and positional effects may produce similar patterns. Averaging attention across heads and layers may also obscure specialized computations. Causal interventions and more tightly controlled contrasts would be needed to establish whether the inferred representations contribute to the models' predictions.

Finally, our experiments examine one English training corpus and one GPT-2-style architecture on a single scale. Generalization to other languages, architectures, and training regimes remains untested. The results establish a computational possibility under this setup; they do not directly determine how humans acquire unlike coordination or resolve competing theories of human grammatical structure.

\makeatletter
\ifacl@finalcopy
\section*{Acknowledgments}
We thank the University of Washington's Hyak Klone HPC
for providing the computational resources used in this work.
\fi
\makeatother


\bibliography{custom}

\appendix

\section{Appendices}
\subsection{Training details}\label{chap:training details}
All models were trained using the same strategy. We used the \texttt{Trainer} class from the \texttt{transformers} package. Training of each model was conducted on one NVIDIA L40 GPU for up to 10 epochs, using the default AdamW optimizer \citep{loshchilovDecoupledWeightDecay2017a} with mixed-precision. The effective batch size was set to 16, the initial learning rate to 5e-5, and the weight decay to 0.01. A complete list of training parameters is provided in Table~\ref{tab:training-params}.

\label{chap:Appendix A - Training Parameters}
\begin{table}[ht]
    \centering
    \begin{tabularx}{\columnwidth}{l *{1}{>{\centering\arraybackslash}X}}
    \toprule
        num\_train\_epochs & 10\\
        per\_device\_train\_batch\_size & 8\\
        gradient\_accumulation\_steps & 2\\
        per\_device\_eval\_batch\_size & 16\\
        learning\_rate & 5e-5\\
        warmup\_steps & 1000\\
        weight\_decay & 0.01\\
        max\_grad\_norm & 1.0\\
        fp16 & True\\
        dataloader\_drop\_last & True\\
        prediction\_loss\_only & True\\
        metric\_for\_best\_model & `eval\_loss'\\
        save\_strategy & `steps'\\
        save\_steps & 1000\\
        save\_total\_limit & 3\\
    \bottomrule
    \end{tabularx}
    \caption{A list of the selected training parameters. Any values not listed, except for logging, are assumed to be the default settings in the \texttt{transformers} package, version 4.56.2.}
    \label{tab:training-params}
\end{table}

\subsection{Examples of evaluation corpora}\label{chap:examples-eval-corpora}
This section presents examples of our evaluation corpora. For sentences in \textit{unlike}, the following examples each represent a different combination pattern in the corpus:

\begin{enumerate}[label=(\arabic*),leftmargin=*]
    \item \textit{You can depend on my assistant and that he will be on time.}
    \item \textit{Voids are a nightmare and initialed by the employees.}
    \item \textit{That was manipulative and designed to keep her in the dark.}
    \item \textit{The phenomenon fell into place organically and with ease.}
    \item \textit{My boss is a perfectionist and demanding.}
    \item \textit{Those leaders insist that they are still in charge and leading the people.}
    \item \textit{I don't know if I should take this exam in class or online.}
    \item \textit{I punched him a lot of times and with all my might.}
    \item \textit{More Americans work out of the house and longer hours.}
    \item \textit{Erosion has left the brick structures crumbling and a clear safety hazard.}
\end{enumerate}

Each sentence in \textit{unlike} has a corresponding counterpart in \textit{alike-from-unlike}, as illustrated by the following examples:

\begin{enumerate}[label=(\arabic*),leftmargin=*]
    \item \textit{You can depend on my assistant and his ability to be on time.}
    \item \textit{Voids are a nightmare and a misery, which is initialed by the employees.}
    \item \textit{That was manipulative and was designed to keep her in the dark.}
    \item \textit{The phenomenon fell in to place organically and easily.}
    \item \textit{My boss is a perfectionist and a demanding person.}
    \item \textit{Those leaders insist that they are still in charge and are leading the people.}
    \item \textit{I don't know if I should take this exam in class or on Canvas.}
    \item \textit{I punched him repeatedly and forcefully.}
    \item \textit{More Americans work out of the house and in longer hours.}
    \item \textit{Erosion has left the brick structures crumbling and dangerous.}
\end{enumerate}

The following examples illustrate the tests used in \textit{unlike-judgment}. For each category, we provide a grammatical sentence together with its corresponding ungrammatical counterpart:

\begin{enumerate}[label=(\arabic*)]
    \item \textbf{Selection Violation}: 
    \textit{You can depend on my assistant and that he will be on time.} 
    *\textit{You can depend on that he will be on time.}

    \item \textbf{Position Effect}: 
    \textit{You can depend on my assistant and that he will be on time.} 
    *\textit{You can depend on that he will be on time and my assistant.}

    \item \textbf{CP as NP Requirement}: 
    \textit{That he will be on time, you can depend on.} 
    *\textit{You can depend on that he will be on time.}

    \item \textbf{Mismatch Restriction}: 
    \textit{Hobbs ended up liking Rhodes and hating Barnes.} 
    *\textit{Hobbs ended up liking Rhodes and to hate Barnes.}

    \item \textbf{Dual Roles}: 
    \textit{St. Peter did reside in Rome.} 
    \textit{St. Peter did die in Rome.} 
    \textit{St. Peter did reside and die in Rome.}

    \item \textbf{Unlike Supercategories}: 
    \textit{Reluctantly and in embarrassment, the white officer released the Black man.} 
    \textit{Reluctantly and embarrassed, the white officer released the Black man.}

    \item \textbf{Deletion}: 
    \textit{I eat meat and at home.} 
    *\textit{Meat and at home do I eat.}
\end{enumerate}

\subsection{Grammaticality judgment results}\label{chap: Grammaticality Judgment Results}
Table~\ref{tab:unlike-gj-detail-result} presents the per-test results of the grammaticality judgment tests.
\begin{table}[ht]
\small
\centering
    \begin{tabularx}{\columnwidth}{l*{3}{>{\raggedright\arraybackslash}X}}
    \toprule
        Test Category & \textit{O} & \textit{All} & \textit{And}\\
    \midrule
selection violation     & \cmark & \cmark & \xmark \\
position effect         & \cmark & \cmark & \cmark \\
CP as NP requirement    & \xmark & \xmark & \xmark \\
CP as NP requirement    & \cmark & \cmark & \cmark \\
verb dependency         & \cmark & \cmark & \cmark \\
mismatch restricted     & \xmark & \xmark & \xmark \\
position effect         & \cmark & \xmark & \cmark \\
position effect         & \cmark & \cmark & \xmark \\
mismatch restricted     & \cmark & \cmark & \cmark \\
mismatch restricted     & \xmark & \xmark & \xmark \\
mismatch restricted     & \xmark & \xmark & \xmark \\
wrong topicalization    & \cmark & \xmark & \cmark \\
deletion                & \cmark & \cmark & \cmark \\
position effect         & \cmark & \cmark & \cmark \\
mismatch restricted     & \cmark & \cmark & \cmark \\
mismatch restricted     & \cmark & \cmark & \cmark \\
dual roles              & \na    & \na    & \na \\
unlike supercategory    & \na    & \na    & \na \\
deletion                & \cmark & \cmark & \cmark \\
deletion                & \cmark & \cmark & \xmark \\
deletion                & \cmark & \cmark & \cmark \\
deletion                & \xmark & \cmark & \cmark \\
    \bottomrule
    \end{tabularx}
    \caption{Results of the three models on each test in \textit{unlike-judgment}. \textit{O} = \textit{Original}; \textit{All} = \textit{Filtered-all}; \textit{And} = \textit{Filtered-and}.}
    \label{tab:unlike-gj-detail-result}
\end{table}

\end{document}